# SPA: Stochastic Probability Adjustment for System Balance of Unsupervised SNNs

Xingyu Yang
*The School of Electronics and Information Technology*
*Sun Yat-sen University*
Guangzhou 510006, China
yangxy266@mail2.sysu.edu.cn

Mingyuan Meng
*The School of Electronics and Information Technology*
*Sun Yat-sen University*
Guangzhou 510006, China
mengmy3@mail.sysu.edu.cn

Shanlin Xiao*
*The School of Electronics and Information Technology*
*Sun Yat-sen University*
Guangzhou 510006, China
xiaoshlin@mail.sysu.edu.cn

Zhiyi Yu*
*The School of Electronics and Information Technology*
*Sun Yat-sen University*
Guangzhou 510006, China
yuzhiyi@mail.sysu.edu.cn

*Abstract*—Spiking neural networks (SNNs) receive widespread attention because of their low-power hardware characteristic and brain-like signal response mechanism, but currently, the performance of SNNs is still behind Artificial Neural Networks (ANNs). We build an information theory-inspired system called Stochastic Probability Adjustment (SPA) system to reduce this gap. The SPA maps the synapses and neurons of SNNs into a probability space where a neuron and all connected pre-synapses are represented by a cluster. The movement of synaptic transmitter between different clusters is modeled as a Brownian-like stochastic process in which the transmitter distribution is adaptive at different firing phases. We experimented with a wide range of existing unsupervised SNN architectures and achieved consistent performance improvements. The improvements in classification accuracy have reached 1.99% and 6.29% on the MNIST and EMNIST datasets respectively.

## I. Introduction

With the development of artificial intelligence, deep learning has achieved good performance in many fields (e.g. recognition, analysis, and inference). However, some scholars have proposed that existing deep learning architectures and methods do not have the potential to reach the human brain [1], and existing deep learning models have high power consumption in hardware implementations, caused by complex neural network architectures and training rules. Even though scholars have raised some models to reduce the power consumption of traditional Artificial Neural Networks (ANNs) [2], this problem has not been ultimately solved. The human brain is an ultra-low-power-consumption system, so there has been a worldwide upsurge of research interest in the models which can perform some of the functions that a human brain does. Consequently, the neural network evolved from the second generation, ANNs, to the third generation, Spiking neural networks (SNNs) [3]. As a core of brain-like intelligence, SNNs have attracted more and more attention.

SNNs consist of input spike generators, computing units (i.e. spiking neurons), connection synapses, and output decoding. SNNs are promising to achieve ultra-low power consumption because each spiking neuron in SNNs works asynchronously in an event-driven manner like the human brain [4-5]. In contrast to ANNs processing continuous signals in deep learning, spiking neurons are similar to biological neurons so that SNNs process discrete signals. However, due to the discrete signal processing in SNNs, some data accuracy will be lost during the conversion from continuity to discreteness [6]. At present, SNNs still cannot keep up with ANNs on image classification tasks. Nevertheless, researchers shouldn't ignore SNNs' potential to produce ultra-high-energy-efficient hardware [7].

For SNNs' training, unlike widely used supervised learning using a loss function to measure the difference between actual output and target output (i.e. label), unsupervised SNNs let neurons adjust their own synaptic weights according to their spiking activities, which is similar to the process of real neurons in the human brain. In addition, researchers should not only focus on the learning methods of a neural network system, because the architecture of a network is also very important for creating a more advanced artificial intelligence system. Some scholars have suggested that the architecture of a neural network itself might be more important than how it learns [8]. Here, by 'architecture' we mean spiking neuron models, synapse models, and connection topology.

Different spiking neuron models have different biological characteristics and computational complexity. Currently, the most widely used spiking neuron models are Spike Response Model (SRM) [33], Leaky Integrate-and-Fire (LIF) model, and Integrate-and-Fire (IF) model [34]. Only a few neuron models were proposed in recent years. Recent researches on spiking neuron models have focused on modeling the features of biological neurons or circuit implementations. For example, the model-nested neuron analysis adopted by [10-11] used a time-domain synaptic efficacy adjustment, which got a similar generalization performance compared to other multi-layer SNNs. Moreover, we can draw on some biological neuroscience content in SNNs researches [9]. The parameters of spiking neuron models such as resting potential, threshold potential, and refractory period can be given by neuroscience [12].

In this paper, based on a stochastic process, we propose a Stochastic Probability Adjustment (SPA) system composed of

This work was partly supported by National Key R&D Program of China under Grant 2017YFA0206200, Grant 2018YFB2202600 and National Nature Science Foundation of China (NSFC) under Grant 61674173, Grant 61834005, and Grant 61902443. * Shanlin Xiao and Zhiyi Yu both are corresponding authors of this paper.

adaptive spiking neuron models and stochastic synapse models with Gaussian-distributed synaptic transmitter. Our system can be used in a wide range of unsupervised SNN frameworks and improve their classification performance. The transmitter's movement is modeled as a Brownian-like stochastic process, and the establishment of synapses is similar to the Ornstein Uhlenbeck Process (OUP) [13]. Through the constraints of the stochastic process, the overall competition of SNNs is reduced, thus improving image classification accuracy. The adaptive adjustments of our SPA are as follows: 1) Adaptive selection of the synapse model. A single synapse has both the characteristics of excitatory and inhibitory synapses. It is no longer mandatory to distinguish between excitement and inhibition. The required type (i.e. excitatory or inhibitory) and distribution of synapses are determined by a Boolean selection of 0 and 1; 2) The reception rate of transmitter self-adapts using Gaussian random process, thereby mitigating the influence of neuron voltage and, 3) The overall stochastic SPA system and it's closed-loop control methods.

## II. RELATED WORK

One of the first unsupervised systems [14] is a simple Fully-Connected (FC) two-layer network using a threshold adaptive mechanism and STDP-based training method. Based on [14], Saunders et al. [15] proposed to adopt Locally-Connected (LC) layers to improve SNN's robustness and training speed, but the improvement in classification accuracy is still marginal. In our previous work [16], we proposed a multi-pathway Inception-like architecture to further improve the robustness and training speed, and this architecture can be extended and used as a multi-layer unsupervised SNN [30]. Panda et al. [17] continued the network architecture of [14] and proposed an Adaptive Synaptic Plasticity (ASP) and an adaptive weight decay mechanism to obtain improved classification accuracy on the MNIST dataset, but their method is limited by slow training convergence. Besides, Spiking Convolutional Neural Networks (SCNNs) [18] combine the multi-layer architecture of CNN and the basic spiking neuron units of SNN to organize unsupervised multi-layer SNN, and get better image classification performance. These multi-layer SCNNs have some common characteristics: Their SNN systems are only used to extract features, and they finally need to utilize an external supervised classifier such as support vector machines (SVM). In this paper, we evaluated our SPA system in the architecture of [14-16] and measured the improvements got by SPA.

Stochastic process models have been widely used in SNNs. Kasabov et al. [19] proposed a probabilistic spiking neuron model where three different probabilistic parameters are used and act together on neural pre-synaptic potential. They also used a probabilistic parameter to quantize transmitter release, and its main role is to affect the membrane potential of a single neuron. In [11], the value of synaptic efficacy function is decided by the sum of amplitude-modulated Gaussian distribution functions and has a supervised adjustment of connection weights through error feedback. Ahmed et al. [20] proposed a general architecture for stochastic SNNs with Bayesian neurons. Also, stochastic SNN architecture can be used in devices, Pagliarini et al. [21] proposed a stochastic SNN architecture which consists of strained magnetic tunnel junction (MTJ) devices and transistors, and tested it in a hand-written dataset MNIST. The benefits of stochastic information theory for neurons have been explained in detail in [22], so we accordingly built a stochastic system, then we analyzed the influences among various signals in SNNs and established a closed-loop influence relationship.

## III. MOTIVATION

### A. Architecture Design

In biological neurons, different relative positions of synapses and soma in a neuron cause different effects of transmitted signals on the neuron, e.g., Excitatory Post-Synaptic Potential (EPSP) or Inhibitory Post-Synaptic Potentials (IPSP). This means that there is no distinction between inhibition and excitation for a neuron itself. As is shown in Fig. 1, we designed that each neuron has both the characteristics of inhibition and excitation, where we use 1 or 0 to represent whether this neuron is in an excitatory or inhibitory state at this time. By determining the types (i.e. excitatory or inhibitory) of all neurons with a distribution matrix containing only 0 and 1, it is easy to build a multi-dimensional complex network architecture that is more similar to the human brain.

### B. Synapse Design

A spiking neuron needs to integrate multiple stimuli to fire a spike signal, and these stimuli come from the linear sum of the effects of excitatory and inhibitory synapses. After the neuron fires a spike, the spike signal is transmitted to downstream neurons through the synapses, and the synaptic transmission is in the form of transmitter. Excluding some indirect intermediate effects, the amount of transmitter received by a post-synapse is jointly affected by multiple independent pre-synapses. Based on this property, we can formulate the synapse transmitter as the probability of Gaussian distribution. This probability indicates the amount of transmitter the post-synaptic neuron receives.

### C. Neuron Design

In the existing neuron models such as Hodgkin-Huxley (HH) model [23] and Izhikevich model [24], the response of neurons is independent during each spike-fired phase, i.e., the neuron's membrane potential responds according to a constraint formula after each spike arrives. For a pattern recognition system, it means that when SNN processes a new sample, this system, except for its connection weights, does not contain any memory for the features of previous samples. The gap between artificial neurons and real biological neurons still exists. The HH model has strong biological plausibility, but it requires excessive computations. Researchers simplified the HH model to a commonly used LIF model, which also discards most of the original image features. These neuron models are limited by the independence of their responding process to information and their separately-processed parameters, i.e., every time the neuron fires a spike, its parameters only 'work' in this spike-fired phase. To reduce the difficulty of spike firing and maintain a balanced SNN system, it is really important to build an efficient spiking neuron model that requires fewer computations and improves parameter utilization in SNNs.

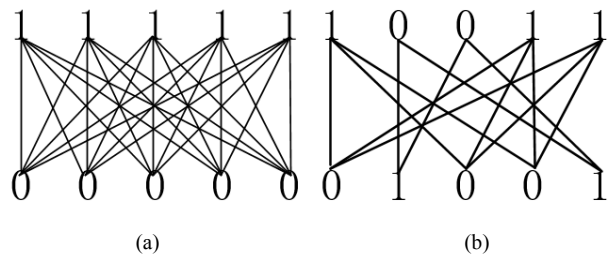

Fig. 1. Examples of general SNN architecture (a) and multi-dimensional SNN architecture with 0/1 distribution matrix (b).

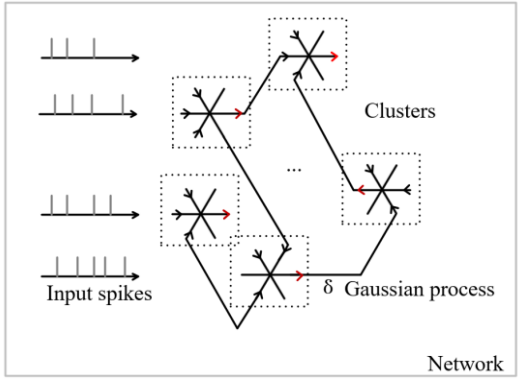

Fig. 2. An illustration of the SNN mapped into probability space $(\Omega, A, P)$ and its synapses clusters.

## IV. METHODS

We mapped SNN frameworks into a probability space in which we used a Brownian-like stochastic process to model synapses transmitting and abstracted spiking neurons as nodes. In this stochastic process, neurons' parameters are influenced by the changes of synapses' parameters to weaken their effects. Neurons adaptively adjust their resting potential through a short-term memory. Our system reduces the competition among neurons and improves the network's performance. The spike excitation (i.e., the total number of fired spikes) is adjusted by an external training algorithm. After mapping a SNN into the probability space, synapse models and neuron models constitute a balanced system. We call it Stochastic Probability Adjustment (SPA) system and it can be used in different SNNs.

Steps:

1) Create clusters with at least one neuron and their connected synapses. The release of synaptic transmitter follows a Gaussian distribution which is the first stochastic process. Each neuron's type (i.e., excitatory or inhibitory) is determined by Boolean value 0 or 1.

2) To distinguish the differences between different clusters, the activities of different clusters are the second stochastic process where we use Brownian-like motion to adaptively adjust variances and movements.

3) The third stochastic process is a statistical OUP, which could predict network evolution.

### A. Synapses Model

As is mentioned in section III, the synaptic signal reception rate $\delta_{i(t)}$ follows a Gaussian distribution, with a certain mean $\mu$ and variance $\sigma$ as follows:

$$\delta_{i(t)} \sim N(\mu, \sigma^2) \quad (1)$$

The range of $\delta_{i(t)}$ is (0,1), with upper and lower limits. If we abstract all synapses and neurons into wires and nodes, ignore the time loss between the nodes, and map the network into a probability space $(\Omega, A, P)$. For a neuron, all connected input synapses are represented as cluster $M$, and the clusters are connected to each other, as shown in Fig. 2.

In the probability set $\Omega$, ignoring the influence of neurons, we can treat the transmission between synapses as a continuous stochastic process, and the transmitter acceptance rate of input synapses follows a Gaussian distribution rule. After the neuron's discharges (i.e. firing), the output random variable starts from a new starting point. The transmitter increments of all clusters in the transmission time interval are independent of each other but only related to the current status. The change of the random variable is continuous in time and it has some properties of Brownian motion. Taking $\delta_{i(t)}$ as Brownian-like motion, we define the difference between the random processes between clusters as follows:

$$\delta(t_i + t) - \delta(t_i) \sim N(\mu, \sigma_i) \quad (2)$$

This $\rho_{i(t)}$ Brownian motion is not a standard Brownian motion, and the mean value $\sigma$ is a non-zero constant. The probability density function is:

$$p(\delta, t_i - t) = \frac{1}{\sqrt{2\pi(t_i-t)}} exp[-\frac{(\delta-\mu)^2}{2(t_i-t)}] \quad (3)$$

Maintain (0,1) as the upper and lower limits of the $\delta_i$ distribution. Since the duration of neuron firing a spike for each cluster is different, we adjust $\sigma_i$ at every phase adaptively, as shown in Fig. 2 Like the Brownian motion, the $\sigma_i$ adjustment increases linearly. The adaptive variance $\sigma_i$ is defined as:

$$\frac{t_{i+1}-t_i}{t_i-t_{i-1}} = \frac{\sigma_i}{\sigma_{i-1}} \quad (4)$$

Variance $\sigma$ has an initial value $\sigma_0$, after the $\sigma$ changes, the $\delta_{i(t)}$ distribution is still limited in range (0,1). If we represent the form of transmitter as conductance, in order to prevent the conductance from generating too small random values and getting a very large voltage when this conductance is the divisor (I/g), we set the lower limit of the $\delta_{i(t)}$ stochastic process to a starting value $\delta_0$. And now the Brownian expectations change as follows:

For $t_1 < t < t_2$, if $\delta_{i(t_1)} = a, \delta_{i(t_2)} = b, \delta_{i(0)} = 0$:

$$E[\delta_{i(t)} | \delta_{i(t_1)} = a, \delta_{i(t_2)} = b]$$
$$= a + (b-a)(t-t_1)(t_2-t_1)^{-1} \quad (5)$$

If there is a hitting value $\vartheta$ with the $\delta_{i(t)}$ first hitting time $T_h$, then:

$$P(T_h \leq t) = \sqrt{\frac{2}{\pi}} \int_{\frac{|\vartheta|}{\sqrt{t}}}^{+\infty} exp(-\frac{x^2}{2})dx = 2(1 - \Phi(\frac{|\vartheta|}{\sqrt{t}})) \quad (6)$$

We define the attenuation form of the transmitter as follows:

$$G_{(t)} = \omega_i exp(-\frac{t_i}{\tau_i}) \quad t_i > 0 \quad (7)$$

All synapses have the same exponential decay form and can be transformed into time differential forms. Note that the connection weight $\omega_i$ represents the reception rate of spike, and it means whether the neuron receives it or not, while $\delta_{i(t)}$ is the receiving proportion of transmitter amount. Excitatory and inhibitory synapses have different decay time constants $\tau_1$ and $\tau_2$. Then, the transmitter between neurons is defined as:

$$g_{i(t)} = \delta_{i(t)} \cdot \omega_i exp(-\frac{t_i}{\tau_i}) \qquad (8)$$

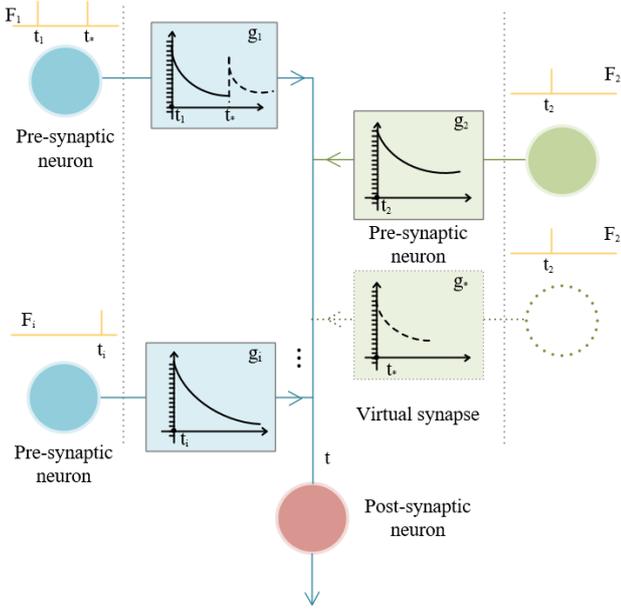

Fig. 3. An illustration of real synapses and virtual synapses on timeline, where the change of post-synaptic potential is caused by the superposition of multiple independent pre-synapses.

As shown in Fig. 3, when a synapse emits more than one spike, the spikes other than the first one are considered to be transmitted by new synapses, ensuring the correspondence between the spikes and synapses and re-establishing the cluster relationship. In fact, the new synapse is virtual just for spike counting. What's more, if there are only fewer synapses emitting spikes in a cluster (e.g. in time-based spike encoding), we will re-divide a cluster and its surroundings into a new cluster.

### B. Neurons Model

We have designed an interpretable, phase-connected spiking neuron model that includes the features of adaptive repolarization [16]. With small polarization retention adjustment, the neuron's response is more stable, and improves training efficiency and neuron's parameter utilization. The response process of the neuron membrane potential to the pre-spikes is similar to OUP. The adjustment of the stochastic process speeds up the convergence of neuron parameters and reduces competition between neurons, making the network more balanced. We use the previous spike-fire phase parameters of the neuron to adjust the initial membrane potential for the next spike-fire phase, i.e. the changes of transmitter. According to the established synapse model, the neuron model is defined as follows:

At first, we define the effect of a single spike $v_{M_i}$ on the spiking neuron's membrane potential as follows:

$$v_{M_i} = \begin{cases} v_0 & t = 0 \\ \Delta v_{t\_f} \int g_{i(t)} \cdot dt & t > 0 \end{cases} \qquad (9)$$

$v_{M_i}$ has an initial value $v_0$ which can be seen as membrane resting potential, and note that the way the neurons receive spike signals and respond to membrane potential is through transmitters. We set the potential difference between the equilibrium potential $E_{eq}$ and the current membrane potential $v_i$ of this neuron when the $i_{th}$ spike arrives as follows:

$$\Delta v_{t\_f} = E_{eq} - v_i \qquad (10)$$

We divide the behavior of neurons into two parts: potential growth before firing and potential reset after firing. The potential growth phase before firing can be a differential form, exponential form, or filter integral form. We define the pre-fire potential growth as follows:

$$v_t = \sum_{i=1}^{M_i} v_{M_i} * \exp\left(-\frac{t-t_i}{\tau_v}\right) \qquad t_i < t \qquad (11)$$

with $\tau_v$ as the time constant of neuron's membrane potential decay. We use the phenomenon of adaptive repolarization proposed by [16] after the neuron's discharge in this design, which can be regarded as an adjustment of the resetting potential after firing, i.e., the potential difference between inside and outside membrane potential. Repolarization is a way to adjust membrane potential to be depolarization or hyperpolarization. When a neuron receives transmitter from its corresponding $M_i$, we use the total amount of received excitatory and inhibitory transmitter as a condition to determine whether a depolarization or hyperpolarization effect happens:

$$\max(\Delta \sum g_e, \Delta \sum g_i) = (1, -1) \qquad (12)$$

$$g_{\sum M_i} = \sum_{i=1}^{M_i} C_i \omega_i \exp(-\frac{t_i}{\tau_i}) = \sum g_e + \sum g_i \qquad (13)$$

$$\Delta \sum g_e = \sum g_{e\_pre} - \sum g_{e\_post}$$

with $\Delta \sum g_e$ and $\Delta \sum g_i$ as the accumulation of different categories of transmitters during two consecutive spike firing phases. There are only two transmitter categories: excitation $g_e$ and inhibition $g_i$ in a cluster $M_i$. We set $\Delta v_a$ as the potential effect of the polarization, and the rule of deciding whether a depolarization or hyperpolarization effect happens is:

$$\Delta v_a = \alpha \cdot \Delta v \cdot \max(\Delta \sum g_e, \Delta \sum g_i) \qquad (14)$$

$$\Delta v = V_{thr} - v_{res} \qquad (15)$$

with a constant $V_{thr}$ as the potential threshold and a variable $v_{res}$ as the resetting potential. It's worthy noting that for a neuron, the value of $v_{res}$ is constant during a period of spike firing and its initial value is $v_0$. The polarization degree after this neuron's discharge is determined by $v_{res}$ and transmitter dominance in previous stage, as shown in Fig.4. And $\alpha$ is an adjustment factor of the degree of polarization as follows:

$$\alpha = \frac{\sum g_e - \sum g_i}{\sum g_e + \sum g_i} \quad (16)$$

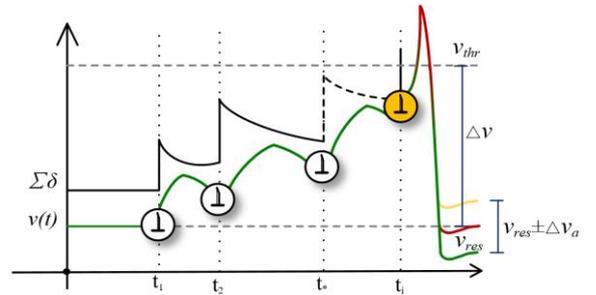

Fig. 4. A neuron fires a spike and adaptively adjusts its resetting voltage with the spikes shown in Fig. 2 as its input.

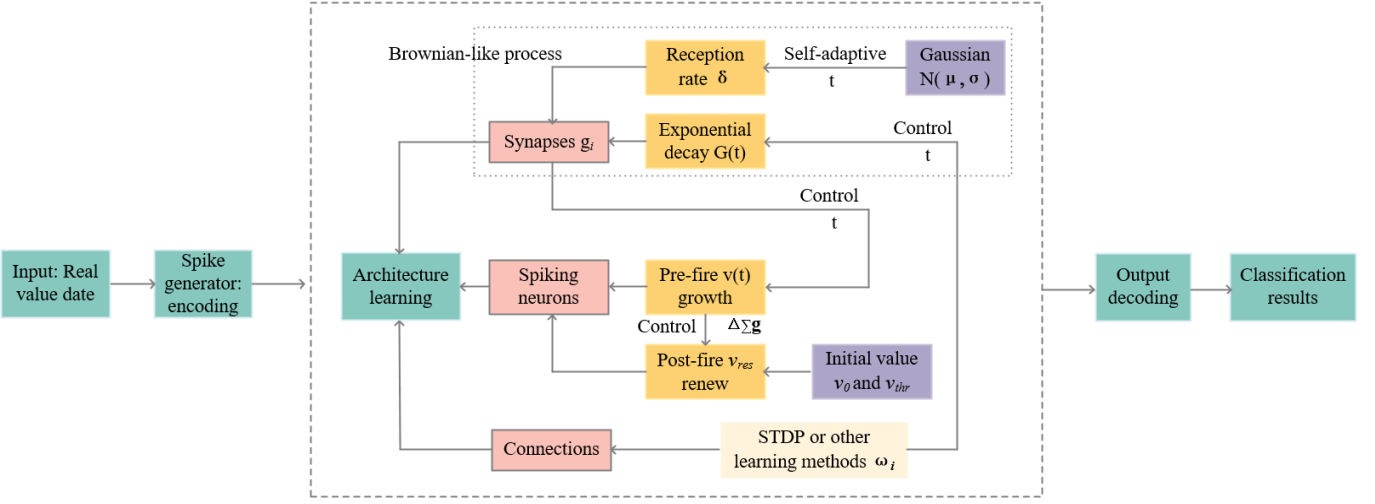

Fig. 5. The overall flow of the SPA system, where the stochastic models are within the dashed line, and the specific stochastic process is within the spoted line. The control process includes self-adaptive adjustment and controlled adjustment.

Here, we make some improvements based on the adaptive repolarization proposed by [16] and get an adaptive factor $\alpha$.

Based on SRM, when a neuron fires a spike and discharges, it will enter a refractory period $\epsilon(\Delta t)$ and be in a resting state after the $\epsilon(\Delta t)$ until it receives the stimulus from the next synapses. The polarization effects after the neuron's discharge are small because the post-fire polarization is local polarization. The adjustment amount in our design is small too (about a few millivolts). We define post-fire potential reset as follows:

$$v_{res} = \hat{v_{res}} + \Delta v_a \quad t > t_f \quad (17)$$

The membrane potential of the neuron model is:

$$v = \begin{cases} v_t & \text{pre\_fire} \\ \epsilon(\Delta t) \cdot v_{res} & \text{post\_fire} \end{cases} \quad (18)$$

The one-dimensional OUP defined by the generalized Wiener random integral is:

$$X_i(t) = C \int_{-\infty}^{t} exp(u - t) \, dB_i(u) \quad i = 1,2,3 \ldots \quad (19)$$

where $B_i(u)$ is a statistically independent generalized Brownian motion process, which can be regarded as transmitter reception rate $\delta_{i(t)}$. Through the analysis of the part $A$, we can regard the change of the transmitter in the neuron as an exponential integral computing core, and now the pre-fire potential changes also take the form of exponential random integration.

In the probability space $(\Omega, A, P)$, we see $v_t$ as an OU-like process which leads us to draw on some of OUP statistical properties to predict SNN performance. We can get the parameters observations $x_{(t)}$ in a period of time calculations:

$$y(t + \Delta t) = f(x_1(t + \Delta t), \ldots, x_n(t + \Delta t)) \quad (20)$$

$x_{(t)}$ can be membrane potential, current, conductance and so on. The forecast value is defined as follows:

$$\bar{Y}(t, \Delta t) = E\{Y(t + \Delta t) | Q_1(x)\} \quad (21)$$

From the statistical properties of OUP, the mean-square error(MSE) corresponding to the forecast value is:

$$\bar{e}(t, \Delta t) = E[Y(t + \Delta t) - \bar{Y}(t, \Delta t)]^2$$
$$= E[Y(t + \Delta t)^2] - E[\bar{Y}(t, \Delta t)^2] \quad (22)$$

Note that the forecast value now is not necessarily the best one with:

$$e(t, \Delta t) \leq \bar{e}(t, \Delta t) \quad (23)$$

We can know the network's working status through MSE prediction.

## V. RESULTS

We tested the proposed method with different unsupervised SNN architectures. The evaluation metrics include classification accuracy, training convergence speed, and training time per sample. We evaluated with MNIST [31] dataset and EMNIST

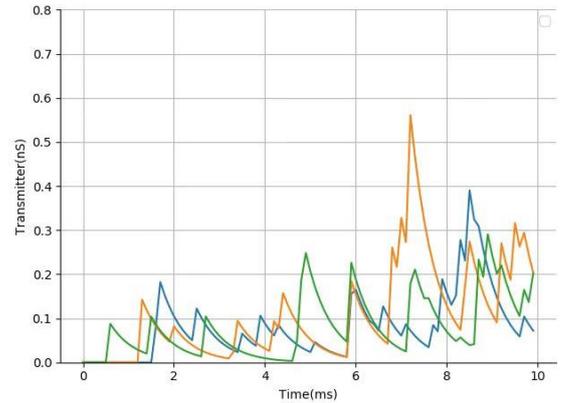

Fig. 6. Three clusters with three neurons are randomly selected and their receiving transmitter distribution are tested. The results are shown as the orange, blue, and green lines.

TABLE I. COMPARISON WITH BASELINE UNSUPERVISED SNN ARCHITECTURES ON THE MNIST

| Paper | Description | Scales (neurons number) | Training samples/convergence | SPA samples | Training accuracy | SPA training results | Testing accuracy | SPA testing results |
|---|---|---|---|---|---|---|---|---|
| Diehl et al. 2015 [14] | 2-layer unsupervised SNN | 6400 | 60000×15times | **60000** | 95.52% | **96.20%** | 95.00% | **95.36%** |
| Saunders et al. 2019 [15] | 2-layer unsupervised LC-SNN | 9000 | 60000 | **60000** | 95.76% | **97.44%** | 95.07% | **95.85%** |
| Meng et al. 2019 [16] | 3-layer unsupervised SNN | 10000 | 60000 | **60000** | 96.02% | **98.01%** | 95.64% | **96.63%** |

[32] dataset in our experiments. We tested the performance of the proposed SPA system on three baseline SNN architectures [14-16] and compared the SPA-enhanced SNNs to state-of-the-art. We got results without changing baseline SNN's original network architecture and hyper-parameters.

All experimented SNNs are trained with a STDP learning rule as follows:

$$\Delta \omega_i = \begin{cases} A^+ \cdot exp(f(x)) & when\ post\ spike \\ -A^- \cdot exp(f(x)') & when\ pre\ spike \end{cases} \quad (24)$$

with $A^+$ and $A^-$ the post-synaptic and pre-synaptic learning rates. The general STDP rule adjusts connection weights according to the range of the time differences. However, in [14-16], their adopted STDP is based on the weight adjustment method of the spike traces and has no concern with time. When a presynaptic neuron or postsynaptic neuron fires, a STDP event occurs. And in [18]:

$$f(x) = -\omega_i, f(x)' = 0 \quad (25)$$

STDP events occur only when a postsynaptic neuron fires. The overall flow of the SPA system is shown in Fig. 5. In our experiments, neurons can be regarded as reservoirs of transmitters. Different clusters have different transmitter cumulative distributions, and every cluster, i.e., pre-synaptic transmitter reception rate, follows a Gaussian distribution. We selected three neurons randomly and plotted the variation of transmitter received by neurons under the control of connection weights $\omega_i$ and reception rates $\delta_i$ in Fig. 6.

*A. MNIST*

MNIST [31] is a hand-written digit dataset including 10000 testing samples and 60000 training samples. All samples are labelled into 10 classes (from '0' to '9'). We applied the proposed SPA to the three baseline unsupervised SNNs and got improvements in training speed and training/testing accuracy. As shown in Table I, our SPA improves the training accuracy of [14], [15], and [16] by 0.68%, 1.68%, and 1.99%, and also improves their testing accuracy by 0.36%, 0.78%, and 0.99% respectively.

What's more, the SPA system has a significant effect on the FC SNN [14]: We only trained it with one pass through of the training set (60000 samples) but obtained a classification accuracy close to the one obtained by the original 15 passes through of the training set. As is shown in Table II, when the neurons number is less than 400, the testing accuracy of [14] can achieve 92.28% with SPA using only 10000 training samples, while the testing accuracy of the original [14] only achieves 87.0% using 60000 training samples. Also, the testing accuracy of [25] using 1000 spiking neurons is only 92.2%. Although the accuracy of [17] reached 96.80%, its training did not converge. SNN's convergence speed is accelerated by adding our SPA. In [18], its result achieved 98.36%, but supervised SVM was used to finish classification. For larger scale SNNs, the training starts to converge after 10000 training samples approximately, and the fluctuation range doesn't exceed 1%. The improvement in convergence speed is relevant to SNN's balance mechanism. The experimental results show that the SPA system composed of stochastic models (i.e. the synapses and neurons) is indeed effective. The establishment of stochastic models reduces the competition among neurons, and the controlled reset voltage with short-term memory effects strengthens the neuron's own responses.

In our experiments, we did not use the adaptive mean value μ but gave it a certain numerical range to avoid the uneven distribution of a cluster's synapses when there are few synapses in this cluster. The change of the mean value may cause some negative effects like making small synapse values ineffective. Since the Wiener random process of synaptic transmitter will weaken its influence on neuron membrane potential, this process reduces competition between presynaptic and postsynaptic neurons. Then, the neuron's autologous response is strengthened through the internal resetting potential adjustment by the transmitter. We also tested the training speed of these SNN architectures before/after using our SPA system and verified whether the proposed mechanism raises the computation training time or not. In the experiments of testing training speed, we used a CPU core for training. The used CPU is Intel XEON 2600 v2/2600. As shown in TABLE III, under the same experimental environment, the effect of SPA on training speed is negligible.

TABLE II. COMPARISON OF TRAINING CONVERGENCE SPEED ON THE MNIST

| Paper | Scales | Origin testing accuracy (using 60000 training samples) | SPA testing accuracy (using 10000 training samples) |
|---|---|---|---|
| Diehl et al. 2015 [14] | 100 | 82.9% | **86.62%** |
| | 400 | 87.0% | **92.28%** |
| She et al. 2019 [25] | 1000 | 92.2% | / |

TABLE III. COMPARISON OF TRAINING TIME ON THE MNIST

| Paper | SPA or not | Training time (s / sample) | Spikes intensity (spikes / sample) |
|---|---|---|---|
| Diehl et al. 2015 [14] | No | 7.26±1 | 6~20 |
| | Yes | **7.48±1** | **4~16** |
| Saunders et al. 2019 [15] | No | 13.68±1 | 12~35 |
| | Yes | **13.24±1** | **11~30** |
| Meng et al. 2019 [16] | No | 13.32±1 | 35~60 |
| | yes | **13.16±1** | **16~50** |

## B. EMNIST

EMNIST [32] dataset is an extension of MNIST from hand-written digits to English letters. We used 120000 training samples and 10000 testing samples from the letter partition of EMNIST, in which all samples are labelled into 26 classes (from 'A' to 'Z'). TABLE IV shows the testing accuracy when our SPA system is used in different SNN architectures. Before adopting the SPA system, the testing accuracy of [14] is too low and the training speed is slow, so we did not perform the EMNIST test of SPA on [14]. As shown in TABLE IV, with the help of our SPA, the training/testing accuracy increases by 4.49%/2.79% in [15], and the training/testing accuracy increases by 6.29%/1.86% in [16]. Besides, Neftci et al. [26] proposed a deep FC supervised SNN using many hidden layers with a spike-based backpropagation as its learning rule. The best testing result got by Neftci et al. [26] is 78.17% on the EMNIST. Also, R. Hu et al. [27] proposed an improved SNN based on [26] and got 85.57% accuracy on the EMNIST at the cost of more computing resources. After adding our SPA system, the comparison of training/testing accuracy among [15], [16], [26], and [27] is shown in Fig. 7. (a), and the training convergence with varying number of training samples is shown in Fig. 7. (b).

TABLE IV. RESULTS COMPARISON ON THE EMNIST

| Paper | Training method | Training accuracy | SPA training accuracy | Testing accuracy | SPA testing accuracy |
|---|---|---|---|---|---|
| Diehl et al. 2015 [14] | Unsupervised STDP | 58.68% | / | low | / |
| Saunders et al. 2019 [15] | Unsupervised STDP | 73.92% | **78.41%** | 67.68% | **70.47%** |
| Meng et al. 2019 [16] | Unsupervised STDP | 83.30% | **89.59%** | 79.86% | **81.72%** |

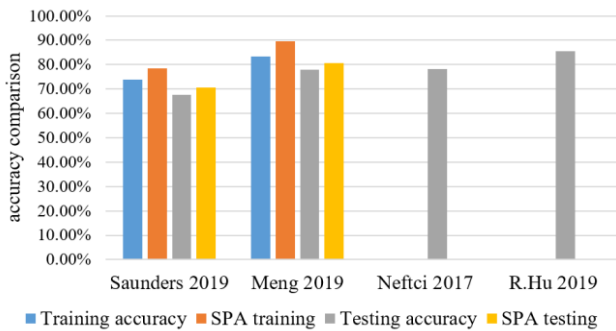

(a)

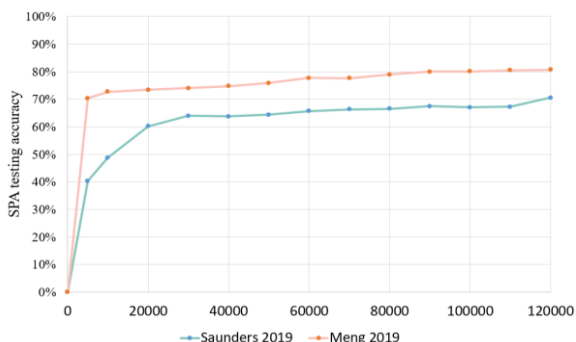

(b)

Fig. 7. Comparison of training/testing accuracy among different SNNs (a); Testing accuracy convergence after adopting SPA on the EMNIST (b).

TABLE V. PARAMETERS SETTINGS IN THE EXPERIMENTS

| Parameter | Description | Value |
|---|---|---|
| $\mu$ | Mean of Gaussian distribution | [0.5,1] |
| $\sigma_0$ | Initial value of variance | 0.25 |
| $\delta_0$ | Lower limit of reception rate | 0.1 |
| $v_{thr}$ | Initial value of threshold voltage | -52mv |
| $v_0$ | Initial value of resetting potential | -65mv |
| $E_{eq}$ | Equilibrium potential | -80mv |
| $\tau_{1/2}$ | Time constant of excitatory /inhibitory synapses | 1/2ms |
| $\tau_v$ | Time constant of membrane potential | 10ms |

Our proposed SPA system can contribute to performance improvements in different SNN architectures. On the one hand, the stochastic probability model is generalizable for different SNN architectures. On the other hand, the optimization methods based on biological characteristics also can facilitate machine learning. The parameter settings in the experiments are shown in TABLE V. Note that some parameters can be found in their corresponding references [14-16] and we didn't list them in TABLE V.

## VI. DISCUSSION

Inspired by neuroscience, we explained the relationship between the characteristics of neuron behaviors and parameters through algorithms, and established a model mapping SNNs into a probability space. Based on the experiments, our research can be further extended in many aspects. The next step may be sparse coding of neuron connections or self-learning of SNN architecture. After obtaining the precise characteristics of transmitter changing, we can try more transmitter modeling other than conductance. Besides, our method is similar to an adaptive generalized Wiener-random process, and the Wiener-process method possibly can be combined with stochastic encoding in digital circuits to improve the performance. Combining neuroscience and computer science is already a key point of machine learning [28], especially for SNNs. In the researches of ANNs, some scholars have begun to explore unsupervised methods, which are considered more advanced [35]. Judged from the results of our experiments, biomimetic methods have indeed produced improved results on a wide range of SNN frameworks, which encourages us to develop more biologically plausible SNNs and implement them on hardware.

## VII. CONCLUSION

We designed a stochastic adjustment system, SPA, for SNNs and designed a stochastic integration core which is similar to OPU for spiking neurons. This mechanism has produced improvements on different unsupervised SNN architectures. The experimental results show that, without more computations required, the maximum improvements of our method in training accuracy are 1.99% and 6.29% on the MNIST and EMNIST, and the improvements in testing accuracy can be up to 0.89% and 2.81% on the MNIST and EMNIST. In addition, we used the statistical properties of the stochastic integration process of the neurons in the mapping model, and used the computing cores in an equivalent differential form when calculating neuron parameters, in order to improve performance without increasing computational cost. Some scholars incorporated a probability-modulated timing method into supervised training and got good performance [29], which suggests the importance of probability models in machine learning. Our experiments also suggest that some characteristics of real biological neurons can facilitate pattern recognition, which seems to be a step approaching brain-like intelligence.